\documentclass[9pt]{IEEEtran}
\pdfoutput=1

\usepackage{times}
\usepackage{helvet}
\usepackage{courier}

\usepackage{fancyvrb}
\fvset{numbers=left,fontsize=\footnotesize}
\DefineShortVerb{\|}
\usepackage{paralist}
\usepackage{graphicx}

\begin{document}

\title{Shadows and headless shadows: a worlds-based, autobiographical approach to reasoning}
\author{
\IEEEauthorblockN{Ladislau B{\"o}l{\"o}ni}\\
\IEEEauthorblockA{
Dept. of Electrical Engineering and Computer Science\\
University of Central Florida\\
Orlando, FL 32816--2450\\
lboloni@eecs.ucf.edu
}
}

\maketitle

\begin{abstract}

Many cognitive systems deploy multiple, closed, individually consistent models which can represent interpretations of the present state of the world, moments in the past, possible futures or alternate versions of reality. While they appear under different names, these structures can be grouped under the general term of {\em worlds}. The Xapagy architecture is a story-oriented cognitive system which relies exclusively on the autobiographical memory implemented as a raw collection of events organized into world-type structures called {\em scenes}. The system performs reasoning by {\em shadowing} current events with events from the autobiography. The shadows are then extrapolated into {\em headless shadows} corresponding to predictions, hidden events or inferred relations. 

\end{abstract}

\section{Introduction}

Many cognitive systems deploy multiple, individually consistent, usually closed models which can represent an interpretation of the present state of the world, a moment in the past, a possible future or an alternate version of reality. These models are often called {\em worlds} or {\em contexts}, although many alternative names exist. For instance, Soar~\cite{lehman1998gentle,lehman2006gentle} dynamically creates structures called {\em substates} whenever it encounters an {\em impasse} in reasoning, and needs new knowledge added to the reasoning pool. In Cyc~\cite{Lenat-1990-Cyc} {\em subtheories} are used to represent alternate versions of reality, for instance, the description of the state of the world in a certain moment in the past (for instance, we can have a microtheory in which Nelson Mandela is still a prisoner). The Polyscheme architecture~\cite{Cassimatis-2004-Integrating} integrates different representations and reasoning algorithms by allowing them to operate over simulated worlds. In \cite{Cassimatis-2009-ReasoningAsSimulation}, the authors show that worlds-based reasoning by simulation can emulate the Davis-Putnam-Logemann-Loveland algorithm and the Gibbs sampling method of probabilistic inference. 
 
The Xapagy cognitive architecture is a recently developed cognitive system, which aspires to mimic the cognitive activities humans use in thinking about stories -- in effect subscribing to the strong-story hypothesis~\cite{Winston-2011-StrongStory}. The system is based on several unusual design decisions. The autobiographical memory is the {\em only} memory model\footnote{When referring to the Xapagy system, we prefer to use the term ``autobiographical memory'' rather than ``episodic memory''. The latter is strongly associated with the work of Tulving~\cite{Tulving-1972-EpisodicAndSemantic}. However, in Tulving's view episodic memory is a ``recently evolved, late-developing [\ldots] past-oriented memory system'' whose ``operations require, but go beyond, the semantic memory system'' \cite{Tulving-2002-FromMindToBrain}. In contrast, the autobiographical memory in Xapagy is not the culmination, but the foundation of all other memory-like behaviors.}. The content of the autobiographical memory is {\em never} extracted into general purpose rules: there is no learning, only a recording of the experiences. While most major cognitive systems implement an episodic/autobiographical memory (see \cite{Nuxoll-2007-EpisodicMemory} for Soar, and \cite{Stracuzzi-2009-IcarusReasoningOverTime} for ICARUS), the importance of the autobiographical memory for Xapagy is more critical. The system has no procedural or skill memory, no rule or production memory, and no concept hierarchy. The content of the working memory (the {\em focus}) can be moved into the long term (autobiographical) memory, but not the other way around. The agent cannot reload a previous experience, nor parts of it. The autobiographical memory influences the behavior of the agent only through the shadowing mechanism.

Xapagy shares with many other cognitive architectures (such as ACT-R~\cite{Anderson-1998-AtomicComponentsOfThought,Anderson-2004-IntegratedTheoryOfMind}) the assumption that acting, witnessing, story following, recall and confabulation are implemented by a common serial mechanism. Together with the other design decisions, this triggers several unexpected implications. The first is the {\em undifferentiated representation of direct and indirect experiences}. The stories exiting from the story bottleneck are recorded together in the autobiographical memory, with no fundamental distinguishing feature. The second implication is the {\em unremarkable self}. The Xapagy agent maintains an internal representation of its cognition (the real-time self), in the form of an instance labeled |"Me"|. However, this instance is not fundamentally different from the instances representing other entities. Together with the inability to recall instances from the memory, this yields another implication, the {\em fragmentation of the self}. As the entity of the self can not be retrieved from memory, only recreated, an agent remembering its own stories will have simultaneously several representations of itself, only one of them marked as its real time self. Thus, {\em every recall of a story creates a new story}. 

Due these unusual design decisions, the Xapagy architecture provides a more fragmented view of reality than other architectures. Situations which in colloquial language appear as an entity going through a series of changes, in Xapagy is represented through several distinct instances, possibly connected by binary relations. The shadowing mechanism aligns the stories recorded in the autobiographic memory with the current focus. The main reasoning model of the Xapagy system operates by extrapolating the shadows of the focus into {\em headless shadows}. These can be seen as predictions (although they can also represent the hidden events, inferred relations or summarizations of the ongoing story). In this sense, Xapagy can be seen to subscribe to the view of the memory-prediction framework of intelligence of Hawkins~\cite{Hawkins-2005-OnIntelligence}. However, in contrast to the Hawkins model, where the memory stores {\em invariant forms}, which denote unique entities, the Xapagy memory only stores raw autobiographical data, distributed over a large number of fragmented instances. 


The objective of this paper is to describe the shadowing / headless shadow reasoning mechanism, with a special focus on how the autobiographical representation and the scene model is a critical ingredient for this reasoning method. 

%
%
\section{Instances, VIs, focus and shadows}

The definition of an instance in Xapagy  is somewhat different from the way this term is used in other intelligent systems. Instead of representing an entity of the real world, it represents an {\em entity of the story, over a time span limited by the additivity of the attributes}. Once an instance acquired an attribute, the attribute remains attached to the instance forever. Things we colloquially call a single entity are represented in Xapagy by several instances. Let us, for instance, consider Little Red Riding Hood (LRRH). There are several versions of the story, ranging from Charles Perrault's (with no happy ending) and the widely known Brothers Grim version, to the countless adaptations and parodies in modern media, including the 2011 movie starring Amanda Seyfried. In Xapagy, these are all different instances, which share the attribute |["LRRH"]|. The Xapagy  system, however, moves a step beyond this. Not only LRRH from the Brothers Grim story and LRRH from the Hollywood movie are represented by different instances, but LRRH the live girl and LRRH the food item in the wolf's belly are also two different instances, as the change can not be represented as an addition of attributes. These instances, can be connected through various {\em relations of identity}: the living and the dead wolf are connected through bodily (somatic) identity, while the real LRRH and the one from the mother's orders is connected by a fictional-future identity relation. 

Events, actions or relations between instances are represented in Xapagy by a {\em verb instance} (VI). There are five structural types of VIs: subject-verb-object, subject-verb, subject-isa-adjective, action-is-adverb and subject-scene-quote. The source of VIs can be outside the agent, in the form of statements written in the Xapi language, which map to one or several VIs. VIs, however, can be also created internally by the processes of summarization, inference of missing actions or relations, recall or confabulation.

The position occupied by worlds or contexts in other cognitive systems, in Xapagy are taken by {\em scenes}. Every instance is assigned to a scene at the moment it is created, and the scene of the instance is not changeable. For most VIs, all the components of the VI are part of the same scene. There are, however, several exceptions. Identity relations might connect instances in different scenes. For quote-type VIs, the quote statement can take place in different scene from the inquit. 

Let us consider an example, where at the end of the story, LRRH, now a grown woman, expresses the fact that she was afraid in Grandma's house facing the wolf. There are two scenes |"StoryEnd"| and |"GrandmasHouse"|. There are also two instances of LRRH in these scenes, which are connected through an identity relation, but otherwise share very little in common. The scenes are connected through a |scene-succession| relation. The Xapi statement will be:

\begin{quote}
\begin{Verbatim}
"LRRH" / says in scene "GrandmasHouse" //
  I / is-a / afraid.  
\end{Verbatim}
\end{quote}

Notice how the word ``I'' in the quote does not refer to the instance of the speaker: it refers to the instance which is identity-connected to the speaker in the scene of the quote.  

\bigskip

The Xapagy equivalent of a working memory is the {\em focus}, a weighted set of recent instances and VIs. In absence of any events, the weights are gradually decreasing. Instances are reinforced when they participate in new VIs. Action VIs are ``pushed out'' from the focus by their successors, while relation VIs stay in the focus as long as their associated instances are in the focus. While in the focus, instances and VIs can acquire new attributes and relations, and they gradually increase their salience in the autobiographic memory. After an instance or VI leaves the focus, {\em it can never return}. 

The instances and VIs from the autobiographical memory affect the current state of the agent by {\em shadowing} the focus. Each instance and VI in the focus is the {\em head} of a an associated instance or VI set called the {\em body} of the shadow. 

The challenge, of course, is how to populate and maintain the shadows such that they reflect the previous experience of the agent with respect to the ongoing story. The system maintains its internal structures using the interaction between a number of {\em activities} which are of an $O(\vert AM \vert)$ or lower complexity, where $\vert AM \vert$ is the size of the autobiographic memory. There are two kinds of activities: {\em spike activities} (SA) and {\em diffusion activities} (DA). SAs are instantaneous operations, executed one at a time. DAs represent gradual changes in the weighted sets; the output depends on the amount of time the diffusion was running. Multiple DAs run in parallel, reciprocally influencing each other. 

In the following we briefly enumerate the SAs and DAs which maintain the shadows. The + or - prefix indicates whether the activities reinforce or weaken the shadow components. The shadow maintenance activities are {\em self-regulating}, encompassing elements of negative feedback as well as resource limitation. 

\begin{compactitem}

\item[(S+)] The addition of an {\em unexpected} instance or VI creates a corresponding empty shadow.

\item[(S+)] The addition of an {\em expected} instance or VI creates a new shadow from the headless shadow which predicted it.

\item[(D-)] In the absence of other factors, all the shadows decay in time. The resources released in this DA are added to the resources of the environment. 

\item[(D+)] Matching the head: instances from memory which match the shadow head will be strengthened in the shadow body. The resources for this DA come from the environment. 

\item[(D+/-)] Consistency: the participation of the VI in a shadow and the participation of its parts in the shadows of the corresponding parts of the shadow head are pulled towards a common value in a resource-neutral way.

\item[(D+/-)] Instance identity sharpening: if an in-memory instance participates in multiple shadows, the strong participations will be gradually reinforced, while the weak participations will be further weakened. The operation is resource neutral for a given memory instance. 

\item[(D+/-)] Non-identity: if a shadow contains instances which are connected through the non-identity relation\footnote{The non-identity relation is explicitly created for distinct instances in the same story line. For example, Achilles is non-identical to the instance of Hector with which it is currently fighting. However, Achilles is {\em not} non-identical with Lancelot.}, the instance with the stronger participation is reinforced while the instance with the weaker participation is weakened. The operation is resource neutral for a given non-identity pair.

\item[(D+)] Identity relation: adds to the shadows instances connected through the identity relation to the shadow head and reinforces those instances.

\end{compactitem}

%
%
\section{Headless shadows}

Headless shadows (HLSs) are collections of related and aligned in-memory VIs which are not paired with any current in-focus VI. The creation and maintenance of HLSs involves three distinct entities: {\em shadow VI relatives (SVRs)}, {\em shadow VI relative interpretations (SVRIs)} and the HLSs themselves. Although for the sake of clarity we describe the creation of these entities in sequential order, in the Xapagy agent these components are maintained by several DAs operating in parallel.

%
%
\subsection{Shadow VI relatives\label{sec:SVR}}

\begin{figure}
\begin{center}
\includegraphics[scale=0.4]{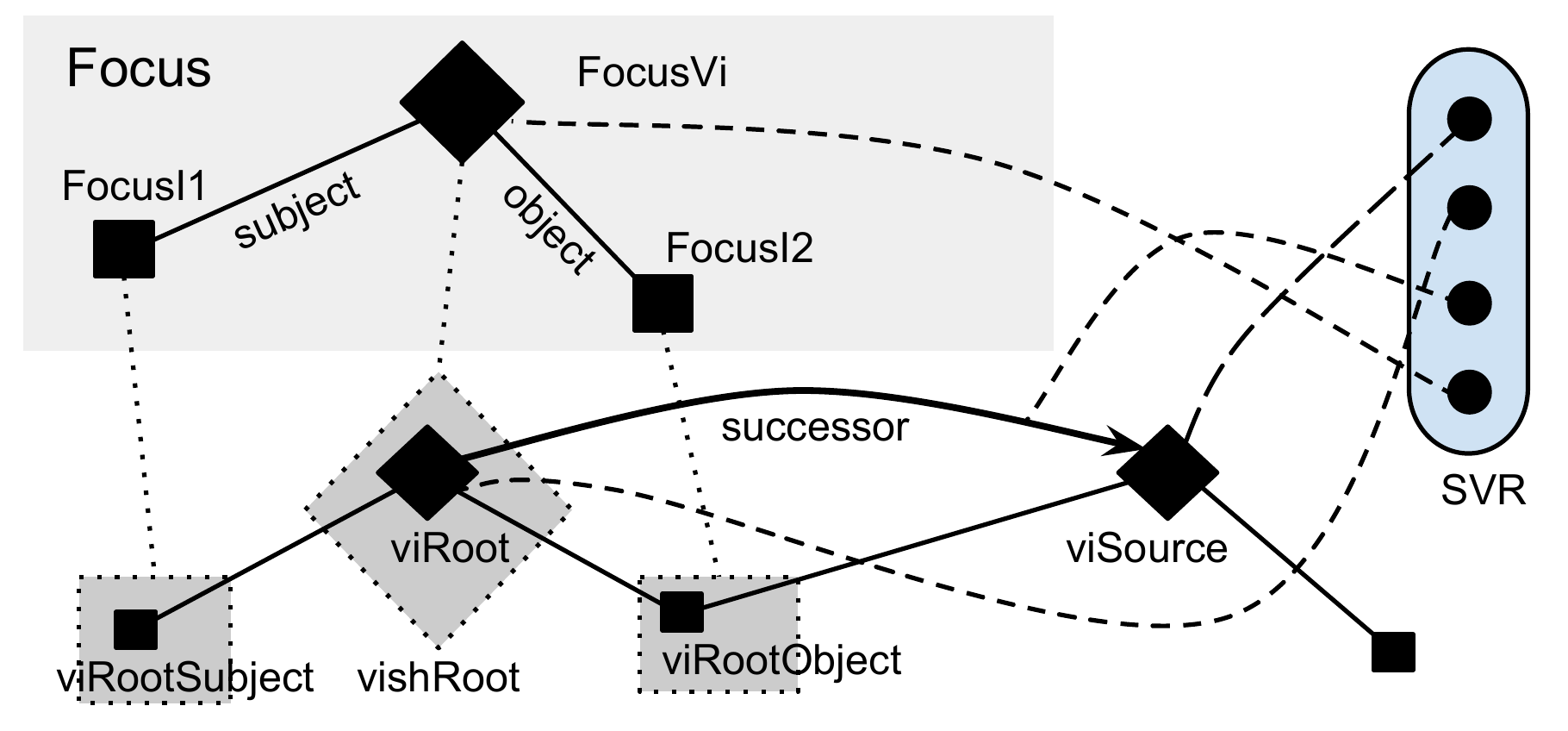} 
\caption{\label{fig:HLS-SVR}The composition of an SVR. VIs are represented as diamonds, instances as squares. Shadows are the same type as the instance they are shadowing.}
\end{center}
\end{figure} 

A shadow VI relative (SVR) is a structure built around a VI which is ``related'' to a VI which is currently in the shadow. Let us consider Figure~\ref{fig:HLS-SVR}. In the focus we have the VI |FocusVI| of type S-V-O, which has its subjects and objects the instances |FocusI1| and |FocusI2| respectively. These elements have their own respective shadows. Let us consider the VI |viRoot| from the shadow of |FocusVI|. This is also of type |S-V-O|, and has its subject and object |viRootSubject| and |viRootObject|, which are part of the shadows of |FocusI1| and |FocusI2| respectively\footnote{These instances can be part of multiple shadows.}. Let us now consider the VI |viSource| which is related to |viRoot| by being connected through a succession relation. The SVR is the quadruplet formed by the |FocusVI|, |viRoot|, |viSource| and the relation type connecting |viRoot| to |viSource|. The latter one is called the {\em type} of the SVR, and it can take nine possible values, all but one arranged in opposing pairs: |IN_SHADOW|, |PREDECESSOR| $\longleftrightarrow$ |SUCCESSOR|, |SUMMARY| $\longleftrightarrow$ |ELABORATION|, |ANSWER| $\longleftrightarrow$ |QUESTION|, |CONTEXT| $\longleftrightarrow$ |CONTEXT_IMPLICATION|.

Intuitively, the |viSource| in the SVR represents a VI which was present when situations similar to the one in the current focus had been encountered. An SVR by itself does not present a prediction with regards to the current focus, because the |viSource| is expressed in terms of in-memory instances, not in-focus instances. In order to find out what kind of prediction does an SVR imply, we must {\em interpret} it.

%
%
\subsection{Shadow VI relative interpretations\label{sec:SVRI}}

\begin{figure}
\begin{center}
\includegraphics[width=\columnwidth]{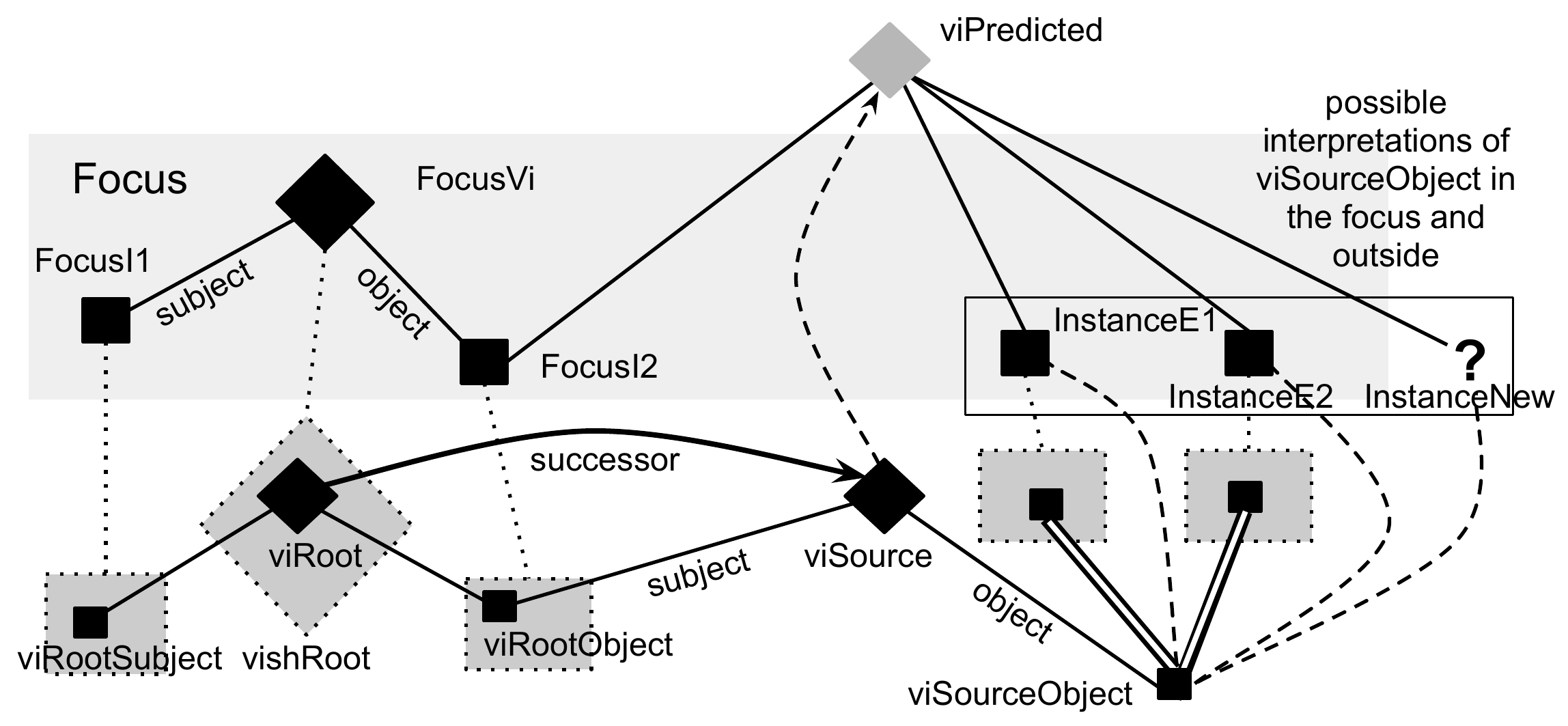} 
\caption{\label{fig:HLS-SVRI} 
Shadow VI relative interpretations.
}
\end{center}
\end{figure} 

Let us take a look at Figure~\ref{fig:HLS-SVRI} which elaborates on Figure~\ref{fig:HLS-SVR}. If |viSource| will be interpreted as a continuation, it will predict the VI |viPredicted| which will also have the format S-V-O. We can make the assumption that the verb in |viPredicted| will be the same as in |viSource|. Finding the subject and the object of |viPredicted| is more complicated: the parts of |viSource| are past instances from the memory which can not be brought back, while the parts of |viPredicted| must be instances in the focus. 

With regards to the subject, we notice that the subject of |viSource| is the same as the object of |viRoot|, which in turn, had been obtained as the shadow of |FocusI2|. We can infer from here than the subject of |viPredicted| will be |FocusI2|. 

The object, however, is more complicated, as it can not be unequivocally inferred from the |viRoot|. Our technique to find an interpretation of |viSourceObject| will be based on ``reverse shadowing'': we look up the shadows in which |viSourceObject| is present, and we consider their heads as candidates with the relative strength of |viSourceObject| in their respective shadows. 

In our case, |viSourceObject| is present in the shadows of |InstanceE1| and |InstanceE2|, these two representing possible interpretations of |viSourceObject|. In addition, there is also a possibility of interpreting |viSourceObject| as an instance which does not yet exist in the focus, |InstanceNew|. 

Putting these considerations together, and denoting the verb of |viSource| with |act|, we have three possible predictions associate with |viPredicted|, which can be described in Xapi as follows:

\begin{quote}
\begin{Verbatim}
The FocusI2 / act / the InstanceE1.
The FocusI2 / act / the InstanceE2.
The FocusI2 / act / an InstanceNew.
\end{Verbatim}
\end{quote}

\noindent where |InstanceNew| is a newly created instance which will be initialized with some of the attributes of |viSourceObject|. These verb instance templates, together with the SVR which are the source of them constitute the SVRIs, weighted by the likelihood of the individual interpretations. 

%
%
\subsection{Headless shadows}

Headless shadows (HLSs) aggregate the support of different types of SVRIs. An HLS is composed of a template for a possibly instantiatable VI and a collection of compatible SVRIs. An SVRI is compatible with a template if the corresponding instance parts are the same and the corresponding concept and verb overlays are ``close''. 

The example used to introduce SVRs and SVRIs was based on a |SUCCESSION| relation, thus we kept referring to them as ``predictions''. The nine different types of SVRIs provide support for or against specific types of HLSs.  Let us consider a case where we see the HLS as a prediction of events to happen next. An SVRI of type |SUCCESSOR| provides evidence that similar events succeeded events in the shadows -- this is a supporting evidence. An SVRI of type |PREDECESSOR| provides evidence that similar events preceded events in the shadow -- which means that they are not successors -- this can be interpreted as a negative evidence. An SVRI of type |IN_SHADOW| means that the given prediction can be mapped back to events which already happened, thus they are not a proper prediction -- again, a negative evidence. An SVRI of type |CONTEXT_IMPLICATION| shows that similar things have happened in similar contexts -- a positive evidence. An SVRI of type |ELABORATION| means that similar things happened when elaborating stories which can be summarized with the same VIs. 

The support of the HLS integrates these evidences into a single number. When an HLS is used to instantiate a new VI, the VI will be created based on the template, while the new shadow will be formed by the |viSource|-s associated with the SVRIs with a positive contribution. After the initial creation of the shadow, this will evolve under the control of the shadow maintenance DAs.

When the HLS is used for a different purpose, the evidences are combined in a different way. For instance, for the inference of a missing action, a |PREDECESSOR| SVRI is a {\em positive} evidence: it can show that we are witnessing the effect of an action which we have not seen. 

%
%
\section{Understanding a children's story}

The objective of the Xapagy system is to mimic the human behavior with respect to stories. In this section, we will use the example of a children's fairytale, Little Red Riding Hood, to illustrate how an agent can be prepared for the understanding of the story, and what types of behaviors we can expect the agent to mimic. The reminder of this section summarizes our ongoing work in this direction. 

%
%
\noindent{\bf Translating LRRH into Xapi.} The first step is to convert the story into a format understandable to the Xapagy system. The English original we used was a hand-written, 500 words long version, which includes a framing device: the narration starts by a little girl Cindy going to bed, and her father is reading her the story of LRRH from the Brothers Grimm fairy tale book. 

\begin{figure*}[t!]
\begin{center}
\includegraphics[scale=0.4]{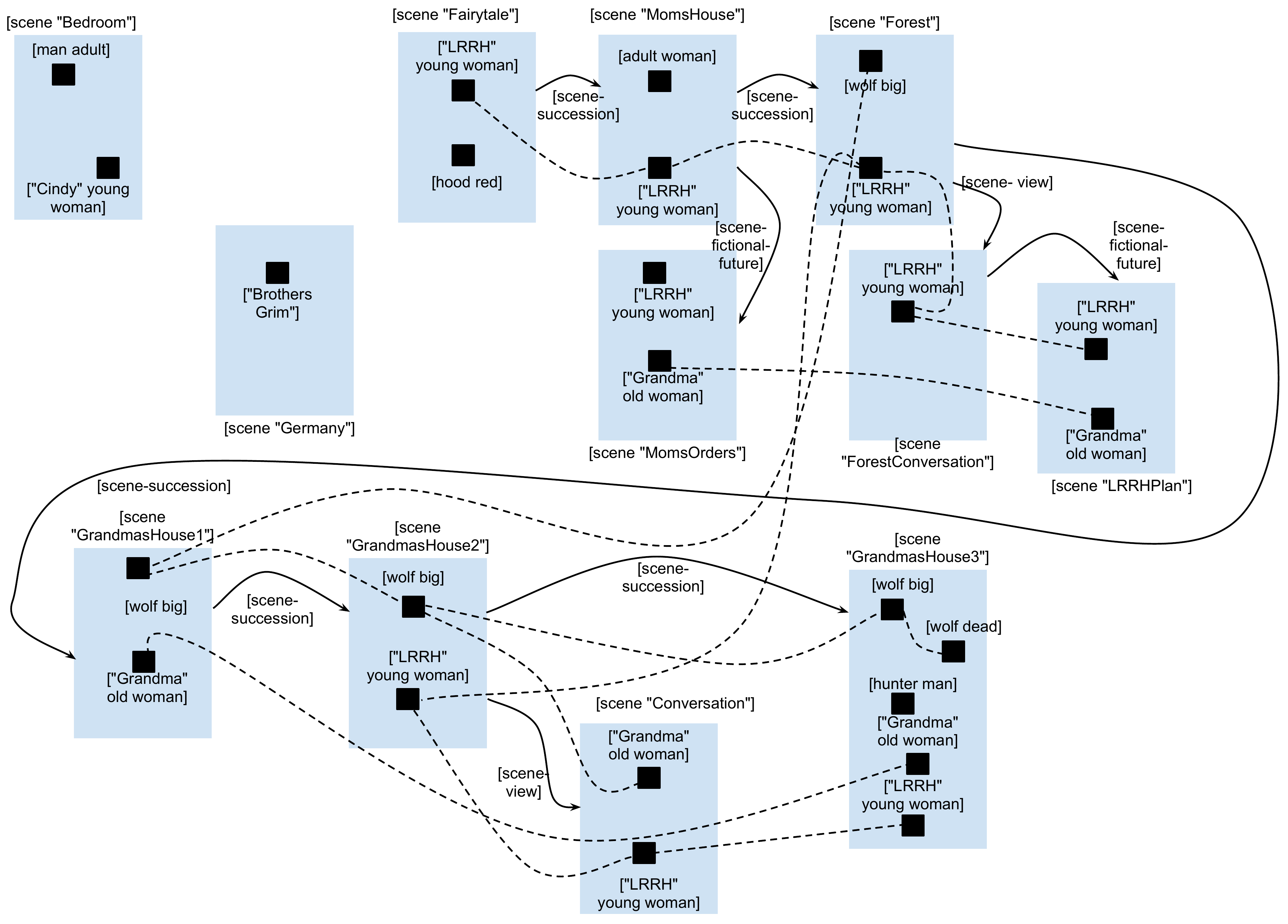} 
\caption{\label{fig:LRRHFull} The scene and identity structure of the story of Little Red Riding Hood written in the
Xapi pidgin and parsed into Xapagy. We ignored some of the entities present in the story (hood, basket and so on), instead focused on the main characters.}
\end{center}
\end{figure*}

The English text had been manually translated to Xapi. The Xapi version is 150 statements long, which, when parsed, is translated into 179 verb instances. Figure~\ref{fig:LRRHFull} illustrates the main set of scenes and instances created through this parsing. The full story is represented through a progression of 12 scenes. Some of these are connected through succession relationships, but we also have separate scenes for the frame, for the dialogs, as well as the representation of LRRH's plans within the dialog. The little girl is represented by 9 distinct instances (including the fictional one in Mom's orders and the planned one when the girl discloses her plans to the wolf). The wolf is represented by 7 instances (including the instance where the wolf impersonates Grandma as well as the alive but sleeping wolf, and the dead wolf shot by the hunter). Translating the story from English to Xapi took about 4 hours of translation work. This number needs to be put in perspective by the fact that this is the first longer story we have translated to Xapi. Xapi is inherently more verbose compared to English, because, in most cases, it requires the explicit specification of the scenes, instances and identity relations.




%
%
\noindent{\bf Extending the concept database and dictionary.} The operation of the Xapagy agent requires a {\em domain library}, consisting of a concept and verb database, and a dictionary which maps words to overlays of concepts or verbs. All Xapagy agents share a core knowledge base (Xapagy Core Domain Library - XCDL) which defines basic concepts such as attributes, scenes, groups, and spatial relations, as well as core verbs such as attribute assignment, instance change and movement in space. About two dozen verbs in XCDL are special purpose {\em metaverbs} which have side effects when activated, and thus carry procedural information. 

The information contained in the Xapagy domain library is very light compared to the domain databases of comparable systems. With the exception of the metaverbs, the definition of the concepts and verbs contain information only about their overlap and impact with other concepts or verbs. The semantics of the concepts and verbs is carried not by the domain library, but by the {\em autobiography}.

To translate LRRH into Xapi, we have reused some of the previously defined domain libraries for topics such as animals, humans, and family relations. To translate the LRRH story into Xapi, we also had to define a number of new concepts, such as ``basket'', ``hood'', ``wine'' and ``swallow''.

%
%

\noindent{\bf Collecting a synthetic autobiography.} Having a translation of a story in Xapi and a corresponding domain library is sufficient for the agent to parse and represent the story. However, the reasoning ability of the Xapagy agent is based on the shadows and headless shadows populated from the autobiography. If there is no relevant information in the autobiography to populate the shadows, no reasoning can take place. There is no shortcut: in Xapagy, we can not replace the autobiography with general purpose rules / scripts / productions, as the shadows only carry particular instances. We can, of course, populate the autobiography with examples which illustrate external rules. 

To achieve the understanding of a story like LRRH, we need to provide the agent with a {\em synthetic autobiography} equivalent to that of a four year old child. While it appears that storing rules and procedures is more efficient than storing a complete autobiography, we find that the size of the human experience is not particularly large compared to the current size of computer storage. If the four old child is witnessing 1 event / second for 16 hours / day (in our opinion, a high estimate), the memory will contain roughly 84 million events, many of them repetitive. This amount of data is computationally quite manageable, as the autobiographic memory, as an unprocessed recording of events, does not suffer from combinatorial explosion. A similar conclusion is reached in~\cite{Laird-2009-YearOfEpisodicMemory}. 


We can, of course, restrict the autobiography to stories relevant to the story at hand. A good rule of thumb is that all the concepts used in the story, as well as their overlap pairs should have supporting stories. 

The upshot is that we do not need to define concepts, only stories in which they appear. Furthermore, the architecture degrades gracefully: the agent can still perform predictions even if some of the concepts introduced do not have associated stories. For instance, we do not need to define what a hunter is: we only need stories in which hunters appear - their actions in those stories will shadow the actions of the hunter in the current story. If the agent had never heard about a hunter, it will still be able to parse the story, but it will not be able to predict that the hunter is carrying a gun, that it will shoot the wolf and so on. 

As an illustration of the nature of the stories we are considering useful for the autobiography for understanding LRRH, here is a collection of stories which we are currently using as a synthetic autobiography for reasoning about LRRH:

\begin{compactitem}

\item Stories where carnivorous animals feed on other animals (lion feeding on a deer, shark feeding on small fish, a Fantastic Mr. Fox eating a chicken).

\item Stories of eating: first person experience that eating something disappears.

\item Stories of conversation: humans take turns saying something. 

\item Stories of question answering: a question is followed by an answer.

\item Stories of following instructions: the actions from a scene are reproduced by the identity linked objects in the other scene. 

\item Stories involving interaction between a grandmother and her granddaughter. 



\end{compactitem}

%
%

\noindent{\bf Reasoning about the story.} The type of reasoning provided by Xapagy is different from reasoning performed by logic based systems. We can not ask the system whether a certain knowledge-base is logically consistent: Xapagy can accept and store logically inconsistent stories. 

Instead, the type of questions a Xapagy agent can answer are along the lines of the informal reasoning people do about stories. Given that we are at a certain point of the story, what do you think it can happen next? What continuations would surprise you? What kind of other events must have happened, which were not explicitly mentioned in the story? If you put itself in place of one of the characters, what would you do next? A Xapagy agent can answer these questions based on the instances and verb instances appearing in the shadows and the different types of headless shadows and their support. 

Let us consider an example. The agent is reading the story of LRRH and the last two Xapi statements read were:

\begin{quote}
\begin{Verbatim}
The wolf / says in scene "Conversation"//
  eyes -- of -- I / sees good / the girl.     
The girl / asks in scene "Conversation"//
   mouth -- of -- "Grandma" / wh is-a / big?
\end{Verbatim}
\end{quote}

What type of predictions will the shadowing architecture make? Examining the continuation-type headless shadows available at this point, we find that the agent predicts two main class of actions. The first class of actions predict the continuation of the conversation:

\begin{quote}
\begin{Verbatim}
Wolf / says in "conversation" / [x]
\end{Verbatim}
\end{quote}

These kind of HLSs are supported by previously seen conversations, where an asking statement had been frequently followed by the other party saying something. It is harder to predict the what the quote |[x]| will be. However, we can make a prediction, that there will be an |ANSWER| relationship between |[x]| and  |mouth -- of -- "Grandma"/ wh is-a / big?|.

Another class of HLSs make the prediction:

\begin{quote}
\begin{Verbatim}
The wolf / eats / the girl.
\end{Verbatim}
\end{quote}

The support for these predictions comes from the scenes involving a carnivorous animal eating another animal. The wolf is shadowed by the tiger, the alligator and ``The Fantastic Mr. Fox'', while the girl is shadowed by the corresponding deer, fish and chicken. Note how the instance of the girl can be shadowed by instances with whom she shares very little.


\if space
Another, somewhat weaker group of prediction appears as follows:

\begin{quote}
\begin{Verbatim}
The girl / gives / the basket.
The wolf / thus takes / the basket.
\end{Verbatim}
\end{quote}

These continuations appear from the previous set of orders by the mother. LRRH is shadowed by instances in previous scenes, in this case the instance from the fictional future of her mother's orders. The wolf's shadow contains the instance of Grandma from these orders, thus the above predictions are made. 
\fi


Let us now consider the way in which the agent uses the pool of predictive HLSs during passive following of a story (for instance, reading, listening or witnessing). Whenever a new action happens, it will be matched against the actions predicted by the headless shadows. If a suitable HLS is found, it will become the shadow of the new VI. The support of the HLS is the level at which the action was ``expected''. The Xapagy architecture defines ``surprise'' as the absolute volume of changes in the shadows and HLSs introduced by a new VI. There are examples of unexpected actions which are not surprises:

\begin{quote}
\begin{Verbatim}
The wolf / sneezes.
\end{Verbatim}
\end{quote}

The agent might have examples of sneezing in its autobiography, however, nothing in the the story had justified the prediction of sneezing. However, once this event actually happens, the corresponding shadow will be established and a prediction will be made for:

\begin{quote}
\begin{Verbatim}
"LRRH" / utters / text "Bless you".
\end{Verbatim}
\end{quote}

The sneezing was unexpected, there was no HLS predicting it. However, it is not a major surprise, as the change in the shadows and HLSs it introduces is minimal and localized. The wolf eating or swallowing LRRH on the other hand, is not unexpected: it is predicted from the stories involving carnivore animals (or from the reproduction of the story, if the agent had previously read it). However, the action, through its impacts, removes LRRH from the focus, and thus the events which involve her future participation will become unsupported. This leads to a major reorganization of the shadows and HLSs. With this definition, the wolf eating LRRH is not unexpected, but it is a surprise. 

\section{Conclusions}

We presented the reasoning technique of the Xapagy architecture. In certain ways the reasoning model is anchored in well known techniques deployed in cognitive systems: it implements the strong story assumption, and uses scenes, a ``world'' type concept to represent the succession of temporal states, fictional entities appearing in conversations and so on. On the other hand, Xapagy is differentiated by its exclusive reliance on the raw autobiographical memory and its fragmentation of real world entities (including the self) into loosely interconnected instances. Accordingly, the reasoning approach, based on shadows and headless shadows, concentrates on aligning the relevant experiences from the autobiography with the current state of the agent and projecting it into predictions about the future and the inference of hidden or unspecified actions or relations in the present.  
\clearpage
\bibliography{Words}

\begin{thebibliography}{10}

\bibitem{Anderson-2004-IntegratedTheoryOfMind}
J.~Anderson, D.~Bothell, M.~Byrne, S.~Douglass, C.~Lebiere, and Y.~Qin.
\newblock An integrated theory of the mind.
\newblock {\em Psychological review}, 111(4):1036, 2004.

\bibitem{Anderson-1998-AtomicComponentsOfThought}
J.~Anderson and C.~Lebiere.
\newblock {\em The atomic components of thought}.
\newblock Lawrence Erlbaum, 1998.

\bibitem{Cassimatis-2009-ReasoningAsSimulation}
N.~Cassimatis, A.~Murugesan, and P.~Bignoli.
\newblock Reasoning as simulation.
\newblock {\em Cognitive processing}, 10(4):343--353, 2009.

\bibitem{Cassimatis-2004-Integrating}
N.~Cassimatis, J.~Trafton, M.~Bugajska, and A.~Schultz.
\newblock Integrating cognition, perception and action through mental
  simulation in robots.
\newblock {\em Robotics and Autonomous Systems}, 49(1-2):13--23, 2004.

\bibitem{Hawkins-2005-OnIntelligence}
J.~Hawkins and S.~Blakeslee.
\newblock {\em On intelligence}.
\newblock Owl Books, 2005.

\bibitem{Laird-2009-YearOfEpisodicMemory}
J.~E. Laird and N.~Derbinsky.
\newblock A year of episodic memory.
\newblock In {\em Proc. of the Workshop on Grand Challenges for Reasoning from
  Experiences}, 2009.

\bibitem{lehman1998gentle}
J.~Lehman, J.~Laird, P.~Rosenbloom, et~al.
\newblock {\em A gentle introduction to {S}oar, an architecture for human
  cognition}, volume~4, pages 211--253.
\newblock MIT Press, 1998.

\bibitem{lehman2006gentle}
J.~Lehman, J.~Laird, P.~Rosenbloom, et~al.
\newblock A gentle introduction to {S}oar, an architecture for human cognition
  - 2006 update.
\newblock URL {\tt
  http://ai.eecs.umich.edu\hspace{0ex}/soar/sitemaker\hspace{0ex}/docs/misc/\hspace{0ex}GentleIntroduction-2006.pdf},
  2006.

\bibitem{Lenat-1990-Cyc}
D.~Lenat, R.~Guha, K.~Pittman, D.~Pratt, and M.~Shepherd.
\newblock Cyc: toward programs with common sense.
\newblock {\em Communications of the ACM}, 33(8):30--49, 1990.

\bibitem{Nuxoll-2007-EpisodicMemory}
A.~Nuxoll and J.~Laird.
\newblock Extending cognitive architecture with episodic memory.
\newblock In {\em Proceedings of the 22nd Natl. Conf. on Artificial
  intelligence - Volume 2}, pages 1560--1565, 2007.

\bibitem{Stracuzzi-2009-IcarusReasoningOverTime}
D.~Stracuzzi, N.~Li, G.~Cleveland, and P.~Langley.
\newblock Representing and reasoning over time in a symbolic cognitive
  architecture.
\newblock In {\em Proc. of the Thirty-First Annual Meeting of the Cognitive
  Science Society}, 2009.

\bibitem{Tulving-1972-EpisodicAndSemantic}
E.~Tulving.
\newblock Episodic and semantic memory.
\newblock {\em Organization of memory}, pages 381--402, 1972.

\bibitem{Tulving-2002-FromMindToBrain}
E.~Tulving.
\newblock Episodic memory: From mind to brain.
\newblock {\em Annual review of psychology}, 53(1):1--25, 2002.

\bibitem{Winston-2011-StrongStory}
P.~H. Winston.
\newblock The strong story hypothesis and the directed perception hypothesis.
\newblock In {\em Proc. of the Advances in Cognitive Systems (ACS-2011)
  Workshop}, 2011.

\end{thebibliography}
\bibliographystyle{abbrv} 

\end{document}